# Using Machine Learning to generate an open-access cropland map from satellite images time series in the Indian Himalayan Region

**Danya Li [1, 2], Joaquin Gajardo [1], Michele Volpi [3], Thijs Defraeye [1,*]**

*[1] Empa, Swiss Federal Laboratories for Materials Science and Technology, Laboratory for Biomimetic Membranes and Textiles, Lerchenfeldstrasse 5, CH-9014 St. Gallen, Switzerland*

*[2] École Polytechnique Fédérale de Lausanne, Lausanne, Switzerland*

*[3] Swiss Data Science Center, ETH Zurich and EPFL, Switzerland*

---

*[*] Corresponding author: thijs.defraeye@empa.ch (T. Defraeye) Empa, Swiss Federal Laboratories for Materials Science and Technology, Laboratory for Biomimetic Membranes and Textiles, Lerchenfeldstrasse 5, CH-9014 St. Gallen, Switzerland*

# Abstract

Crop maps are crucial for agricultural monitoring and food management and can additionally support domain-specific applications, such as setting cold supply chain infrastructure in developing countries. Machine learning (ML) models, combined with freely-available satellite imagery, can be used to produce cost-effective and high spatial-resolution crop maps. However, accessing ground truth data for supervised learning is especially challenging in developing countries due to factors such as smallholding and fragmented geography, which often results in a lack of crop type maps or even reliable cropland maps. Our area of interest for this study lies in Himachal Pradesh, India, where we aim at producing an open-access binary cropland map at 10-meter resolution for the Kullu, Shimla, and Mandi districts. To this end, we developed an ML pipeline that relies on Sentinel-2 satellite images time series. We investigated two pixel-based supervised classifiers, support vector machines (SVM) and random forest (RF), which are used to classify per-pixel time series for binary cropland mapping. The ground truth data used for training, validation and testing was manually annotated from a combination of field survey reference points and visual interpretation of very high resolution (VHR) imagery. We trained and validated the models via spatial cross-validation to account for local spatial autocorrelation and selected the RF model due to overall robustness and lower computational cost. We tested the generalization capability of the chosen model at the pixel level by computing the accuracy, recall, precision, and F1-score on hold-out test sets of each district, achieving an average accuracy for the RF (our best model) of 87%. We used this model to generate a cropland map for three districts of Himachal Pradesh, spanning 14,600 km$^2$, which improves the resolution and quality of existing public maps.

## Keywords

# 1 Introduction

Land cover, cropland, and crop type maps can provide crucial information for agricultural and food security analyses and decisions. These maps are also of interest to microscale solar-powered cold storage facilities providers for smallholder farmers and farmer producer organizations. They can help these stakeholders to better decide where to locate their facilities to create the highest impact in preserving the produced crops.

Two global open-access products, the Copernicus land cover map of 2015 [1] and the Global Food Security-support Analysis Data (GFSAD) cropland map of 2017 [2], only have coarse 100-meter and 30-meter resolution, respectively. High-resolution remote sensing imagery, such as those provided by the Sentinel-2 twin satellites from the European Space Agency (ESA), provides an interesting alternative for creating cost-effective and more reliable regional land cover and agricultural maps, especially in smallholding regions. Sentinel-2 images are accessible for free in the Copernicus Open Access Hub [3] and cover the entire globe up to 10 m ground sampling distance (GSD) resolution every five days. The high spatial, spectral, and temporal resolution of these images provides a rich data source that allows for better pattern recognition and data mining by models aimed to generate the aforementioned types of maps.

ML methods have been successfully applied together with remote sensing imagery on a wide range of agricultural applications, including cropland and crop type classification [4], crop yield estimation [5], and precision agriculture [6], and used to generate thematic maps as a result [1], [2], [7]. The different techniques employed for these applications range from traditional ML models to modern Deep Learning (DL) models, which are reviewed in more detail in the following section. Nevertheless, generalizing these approaches to remote regions in developing countries that are dominated by smallholder farming remains a difficult task, primarily due to the lack of ground truth data necessary for training and validating the models.

We aim to help increase smallholders' access to cooling in India by providing cooling service providers with up-to-date information about the regional distribution of agricultural activity. To this purpose, we generate a binary 10 m resolution cropland map for the year 2020, spanning over three districts of Himachal Pradesh, depicted in Figure 1. We developed an end-to-end ML pipeline for the acquisition and processing of raw Sentinel-2 satellite images time series, feature engineering, ML model selection, and its testing and application for inference over a large-scale area. The output map is made openly-available in Google Earth Engine for visualization and download (https://code.earthengine.google.com/2f496bca8eeed758b4e61576428815a9). Additionally, the map is included as a data layer on an open-source web application designed to facilitate the site selection of new cooling rooms (https://yourvcca-maps.users.earthengine.app/view/yourvcca-map-india) [8].

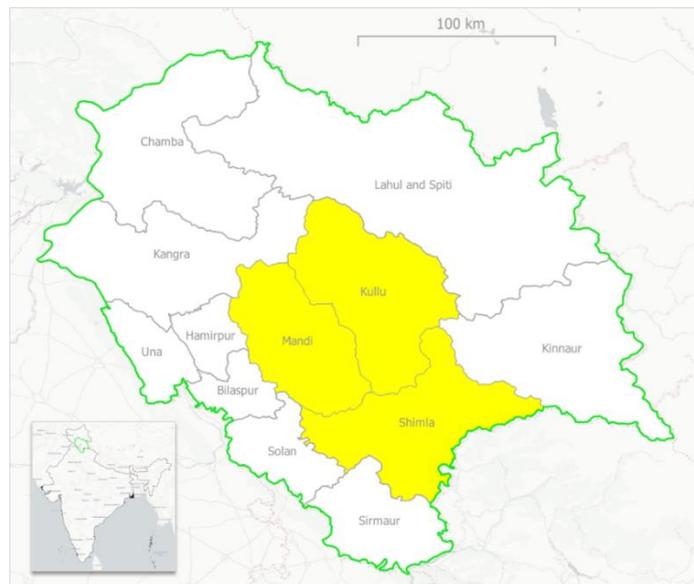

*Figure 1. Location of our study area in Himachal Pradesh (in green), India: the districts of Kullu, Mandi, and Shimla (in yellow).*

## 2   Related work

Traditional approaches for cropland and crop type classification have been studied since several decades, in which including manually crafted features is a key step. In previous studies, the multi-spectral bands in a time series of satellite images are either utilized as features directly [7], [9]–[11] or fitted to some distribution or parametric model to derive informative features for classification [12]. One can also leverage related expert knowledge to build temporal features for cropland mapping [13]. Many studies use vegetation indices computed from multi-spectral bands, and features derived from them, to help the models differentiate cropland from other land cover types [7], [14]–[16]. According to [11], [14], [16], the most significant index to discern cropland from other land cover in satellite images is the normalized difference vegetation index (NDVI) [17]. Other important indices are the green chlorophyll vegetation index (GCVI) [14], brightness, and the normalized difference water index (NDWI) [11]. Statistical [9], [11] and textural features [11], [18] derived from the images are also often considered. The work of [19] designed spectral-temporal features corresponding to the reflectance values observed at the key stages characterizing the crop development cycle. Additional data, such as topographical information [7] and thermal data [15], can also be incorporated to increase the accuracy of cropland or crop type classification.

Among the supervised ML models used for cropland classification, random forest (RF) [7], [11], [16], [18], [20] and support vector machines (SVM) [13], [16], [18] are still the preferred choices, especially for large-scale applications and when ground truth annotations are scarce. However, the generalization performance of these models highly depends on the robustness and appropriateness of the hand-crafted features, hence careful feature engineering is required when using these methods. One study [7] produced a continental-scale cropland map for North America by combining a RF and another segmentation approach. The global GFSAD cropland map was also produced using two pixel-based classifiers, a RF and a SVM, combined with a recursive and hierarchical object-based classifier [2]. Another global map, the Copernicus land cover map, combined the use of a RF classifier and other expert rules [1]. In spite of being readily available, both of these maps have the limitation of being of low spatial resolution, which makes them unreliable in smallholder contexts, where the farm size is often below one hectare.

In recent years, deep learning approaches have also been successfully applied in crop mapping [4], [14], [21], [22]. These methods offer the possibility of avoiding manual feature engineering by directly learning optimal data-dependent feature representations for the task at hand. Most of these approaches rely on neural network architectures that are specialized for explicit time series modeling, e.g., recurrent neural networks. However, these models usually come at the expense of requiring a considerable amount of high-quality ground truth data for training, and computing power. In developing countries and remote areas, the access to ground truth data is often the main barrier for the application of these methodologies.

Overall, ML approaches have been well studied for performing cropland classification with remote sensing imagery. The availability of ground truth data is often a decisive factor for researchers to decide in favor for a specific methodology. Given the scarcity of cropland references points in our study area, we chose to test traditional machine learning methods with explicit feature engineering of the time series of satellite images.

## 3   Materials and Methods

### 3.1   Machine learning pipeline overview

Our data processing pipeline for cropland mapping with ML is summarized in Figure 2. The pipeline includes several steps from the acquisition of all satellite images over a year, preprocessing, feature extraction, training and validation of a cropland classifier, to the generation of a cropland mask for the study region. The code is written in Python and openly accessible via GitHub: https://github.com/YourVirtualColdChainAssistant/MLsat-cropland-himachal.

First, we obtain a dense time series of Sentinel-2 satellite images for the entire year 2020 over our study region and preprocess it. The preprocessing block is done per-tile and it consists of atmospheric correction, masking out clouds, handling missing values, data normalization, and producing a fixed-length time series by temporal sampling and aggregation of single images. Next, we compute the NDVI and extract several temporal, statistical, and spatial features to generate a feature vector of constant dimensionality for each tile and aggregate them. The resulting pixel-wise feature vector is matched to the available ground truth, i.e., cropland or non-cropland, and it is used to train, validate and test two models for cropland classification, a RF and an SVM. Finally, the selected model is applied on three non-adjacent (i.e. spatially independent) regions to assess generalization accuracy, and used to generate the final cropland mask of the study region.

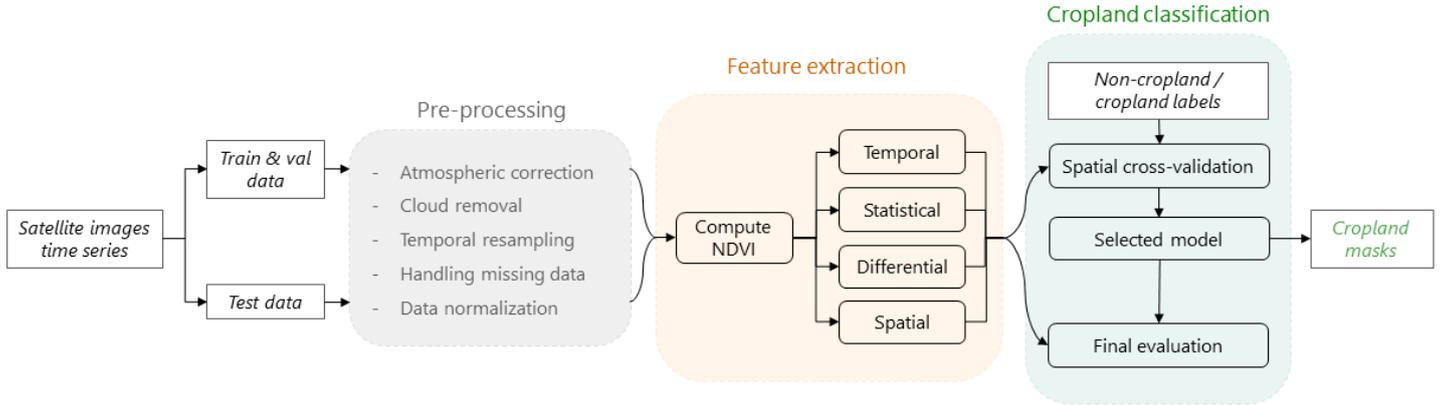

*Figure 2. The pipeline of our pixel-based cropland classification model. It comprises three main blocks: data preprocessing, feature extraction, and cropland classification.*

## 3.2 Satellite imagery

The open-access Copernicus Sentinel-2 satellite images are used in this work due to their accessibility and good temporal, spatial, and spectral resolution. Two Sentinel-2 products are available for users on the Copernicus Open Access Hub [3]: Level-1C images, providing Top-Of-Atmosphere (TOA) reflectance as measured at the top of the atmosphere, and Level-2A provides Bottom-Of-Atmosphere (BOA) reflectance, thus corrected from atmospheric distortions. In the pipeline, Level-2A images are downloaded whenever available; otherwise, they are derived from Level-1C products with the Sentinel-2 Toolbox software `Sen2Cor` [23]. This processor not only performs the atmospheric correction but also provides a Scene Classification layer (SCL) with additional layers such as cloud cover, shadows, and snow. We downloaded all Level-1C images of the year 2020 for the four tiles that include our study region, given that Level-2A images were not available. This is done by querying from the Copernicus Open Access Hub using the `sentinelsat` Python package [24]. Figure 3 shows examples of a given tile across 2020.

With an average revisit time of five days [25], there are generally about 70 images per year for each tile. The number can increase when overlaps between adjacent satellite orbits happen. The Multispectral Instrument (MSI) on board of the Sentinel-2 satellites measures reflectance data on 13 spectral bands, from the visible and near-infrared to shortwave infrared (SWIR), at different spatial resolutions: four at 10 m, six at 20 m, and three at 60 m GSD [25]. Here we consider only the 10 m resolution bands, i.e., the B02 (blue), B03 (green), B04 (red), and B08 (near-infrared, or NIR). For Sentinel-2 products, the tiles are 100x100 $km^2$ orthorectified images. Partially covered tiles may occur if the tile is at the edge or top/bottom of the satellite's swath [25], as illustrated in Figure 3.

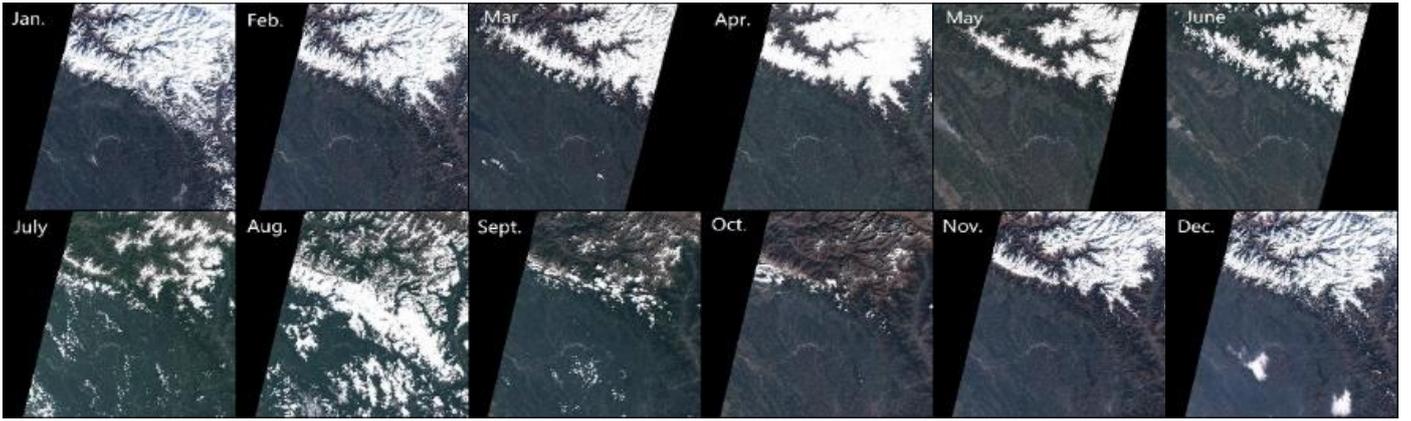

*Figure 3. True color images of remote sensing examples on tile 43SFR over 2020. Each image has a scale of 100x100 km². We can see the earth's surface evolves accordingly.*

### 3.3 Data preprocessing

As introduced in Section 3.1, we take the following data preprocessing steps: 1) atmospheric correction, 2) cloud removal, 3) regular temporal sampling, 4) handling missing data, and 5) data normalization. The steps are detailed in the following and examples about the effects of each step are visualized in Figure 4.

1) **Atmospheric correction**. After downloading the images, we perform atmospheric correction on them with the `Sen2Cor` software [23].

2) **Cloud removal.** We used the SCL data from `Sen2Cor` output to mask out pixels categorized as low / medium / high probability clouds, thin cirrus, and cloud shadows.

3) **Regular temporal sampling**. We made a new time series composite by weeks to keep the time series length constant, which is critical for having feature vectors of fixed dimensionality across different tiles. This was done by taking the first cloud-free time step in a week. Thus, in 2020, for any given tile, the time series of images is of length 53.

4) **Handling missing data**. At a given timestamp per-pixel missing values may occur after cloud removal, or due to the satellite orbit, as exemplified in Figure 3. In both cases, missing values are filled by linear interpolation in each band along the temporal dimension. Other options of data imputation were evaluated and presented in Section 10.3. Any pixel with data completely missing during the year, is removed.

5) **Data normalization**. We considered four different methods to normalize our data, given this is an important preprocessing step for many ML models: i) scale between [0, 1] (referred as `as_float`); ii) divide the digital numbers (DN) by 10000 to obtain the true reflectance units [25] (`as_reflectance`); iii) min-max scaling of each feature on the input vector to [0, 1] (`standardize`); iv) per-feature normalization to mean zero and variance one (`normalize`). The main difference between the first and third options is that `as_float` scales according to the theoretical maximum and minimum values, given the data type registration of the image, whereas `standardize` scales are based on the observed ones in addition to being standardize per feature. Another key difference between the methods, is that i) and ii) are performed globally preserving temporal trends and variations, while iii) and iv) are performed per feature, de-facto de-trending time series. We chose the second option, `as_reflectance`, following other work using Sentinel-2 images and experimentally validated the choice with experiments presented in Section 10.3.

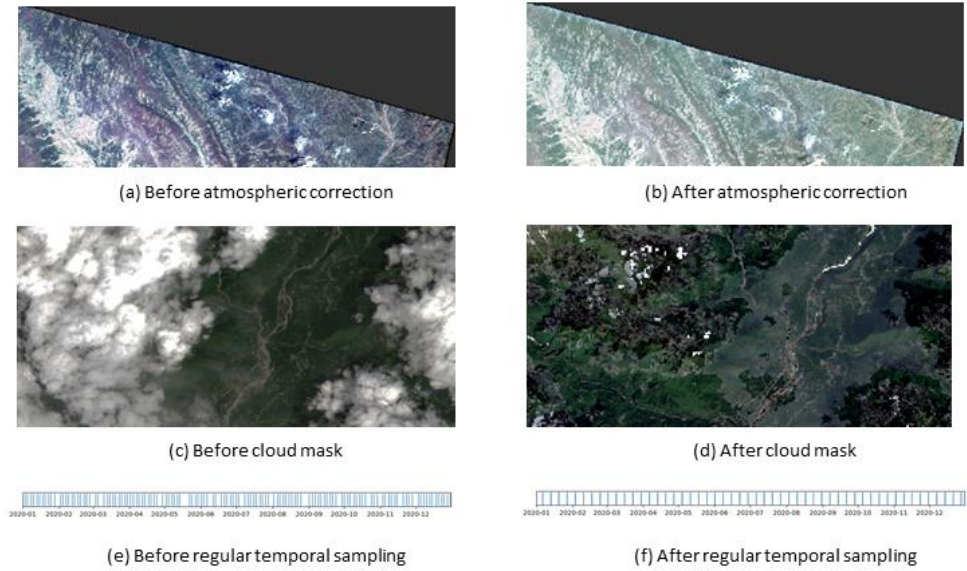

(a) Before atmospheric correction

(b) After atmospheric correction

(c) Before cloud mask

(d) After cloud mask

(e) Before regular temporal sampling

(f) After regular temporal sampling

*Figure 4. Examples of the effect of the data preprocessing steps, including atmospheric correction, cloud mask (both for a single time step), and the regular temporal sampling.*

## 3.4 Feature engineering

Traditional ML models often require manual feature extraction in order to perform acceptably. Feature engineering consists of the process of defining new informative features, based on expert knowledge, and appending them to the feature vector. We aim to generate a feature vector of a fixed dimensionality for each pixel from the satellite images. The computed features can be categorized into four groups: i) temporal features, ii) statistical features, iii) differential features, and iv) spatial features. It is worth noting that all these features are derived from the original satellite images bands, thus no auxiliary data is used for feature engineering in this study. A summary of the computed features is given in Table 1, and details are given in the following.

As previously described, at each weekly time step, the NDVI has been computed and added as an additional feature to the RGB and NIR bands. As previously mentioned in Section 3.3, the data have been temporally aggregated to a weekly composite. The temporal features correspond to taking the values of each band every four weeks, i.e., 70 features equivalent to the values of 14 timesteps for each of the five bands. These temporal features are helpful to characterize the patterns in the temporal vegetation profile in absolute values and in a coarser temporal scale. Additionally, per-pixel statistical features are able to capture significant characteristics during the growing process. The statistics we compute are the average, maximum value, and the standard deviation of all bands along the full temporal dimension of our weekly composite, so they account for 15 additional features. The differential features are the differences between two successive values for a given band in the temporal dimension. This difference captures temporal changes from a relative perspective. Only NDVI is considered for this type of feature, and the weekly differences in values are computed, equivalent to 52 additional features. Spatial features provide information about the surrounding pixels. For every pixel and at each week, we take its contiguous pixels and compute the mean and standard deviation on each band, thus accounting for 530 additional features. Therefore, we generate a feature vector of 667 features in total for each pixel.

*Table 1. Feature engineering summary. There is a total of 667 features for each valid pixel, categorized into four features types. Five bands are considered, RGB, NIR and the derived NDVI. Temporal features are the per-band maximum pixel value every four weeks. Statistical features are the overall per-band average, maximum and standard deviation of each pixel in the entire year. Differential features are the weekly differences of NDVI values (to the previous week). Finally, spatial features for a given pixel are the weekly per-band mean and standard deviation of all its contiguous pixels.*

| Feature type | Temporal | Statistical | Differential | Spatial |
|---|---|---|---|---|
| *Features per band* | 1 | 3 | 1 | 2 |
| *Bands* | 5 | 5 | 1 | 5 |
| *Timesteps* | 14 | - | 52 | 53 |
| *Total* | 70 | 15 | 52 | 530 |

## 3.5 Labeled data

In the context of the funding project, we commissioned a socio-economic ground survey and obtained the GPS locations of apple farmers in Himachal Pradesh. These survey GPS points served as a reference for inspecting the area with VHR base maps and drawing and annotating cropland and non-cropland examples. All the example polygons are drawn in the free and open-source QGIS software [26], using Google Satellite and Bing Satellite base maps as a reference, together with the free Google Earth Pro software.

The polygons labeled as non-cropland include forests, built-up areas, rivers, grassland, etc. Referring to the labeling procedure in [27], we use the strategy to first train a naive model with a few initial training polygons and then gradually create more samples by analyzing the predicted map and identifying the misclassified points. In this way, we maximize the utility of each label, employing few human-in-loop iterations.

As shown in Figure 5, our labels are distributed between three districts. We divided the data into a big training and validation dataset for Kullu (Figure 5a), including 80% of our labels, and left the remaining labels for testing purposes. One test region is near the training set, and the rest are in the two other districts. The purpose is to test the models in near and far-away areas to demonstrate their generalization ability. In Table 2, we present the pixel count of the dataset, showing that for all sets, the amount of samples per-class is roughly balanced.

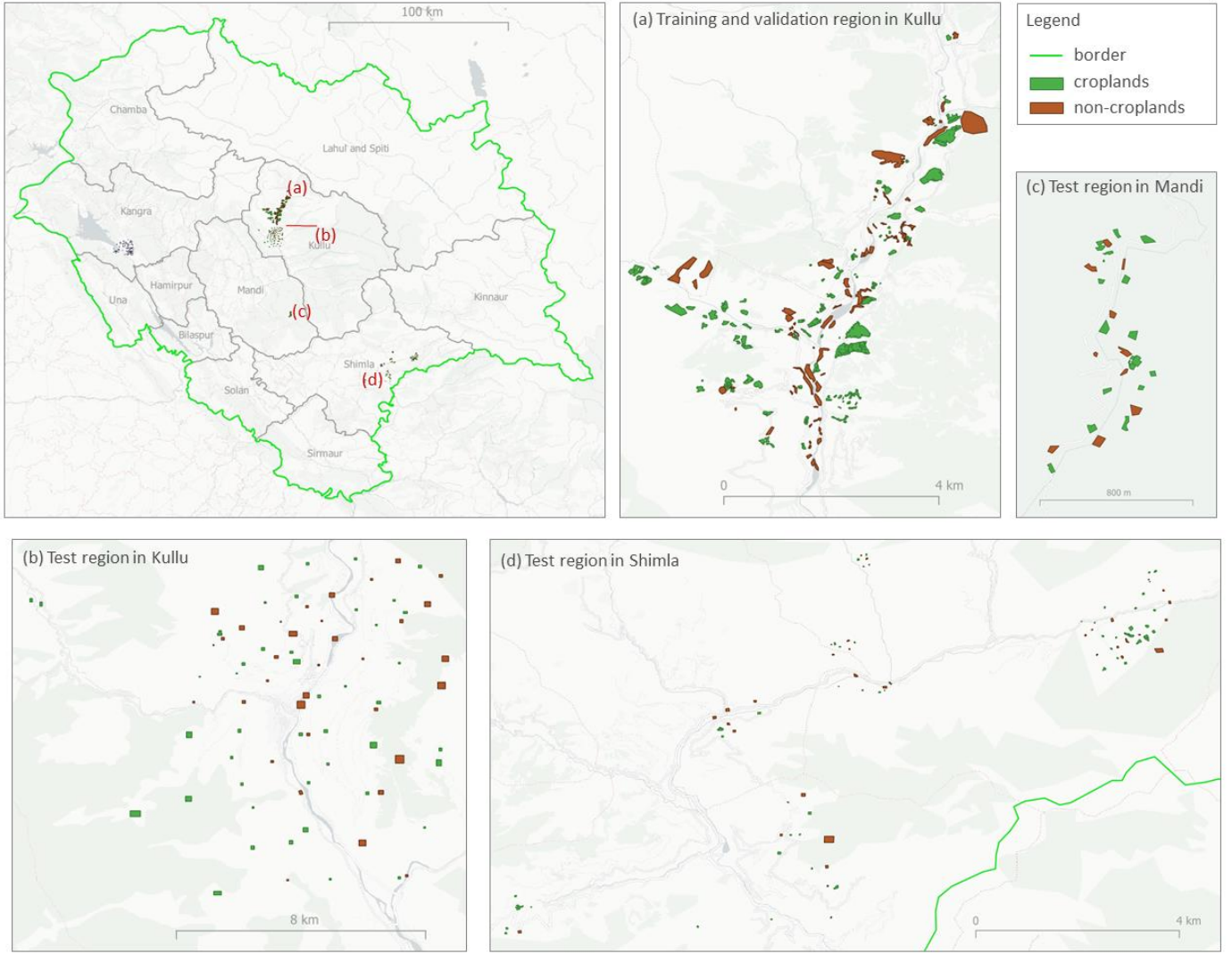

*Figure 5. Labels distribution in our study region. There is one large training and validation region (a) in Kullu, and three small testing regions (b, c, d) in Kullu, Mandi, and Shimla.*

*Table 2. Dataset summary of cropland and non-cropland labels for the training and test regions. The number of pixels represents the total pixel count that intersect a polygon of either class.*

| Class / District | Kullu [train] | | Kullu [test] | | Mandi [test] | | Shimla [test] | |
|---|---|---|---|---|---|---|---|---|
| | Pixel count | % | Pixel count | % | Pixel count | % | Pixel count | % |
| Croplands | 54724 | 52.28% | 6548 | 47.36% | 1198 | 60.97% | 4642 | 47.95% |
| Non-croplands | 49944 | 47.72% | 7277 | 52.64% | 767 | 39.03% | 5038 | 52.05% |
| Total | 104668 | 100% | 13825 | 100% | 1965 | 100% | 9680 | 100% |

Additionally, we also visually assess the quality of our labels by plotting the per-class smoothed average NDVI time profiles of the weekly time series composite in Figure 6. Similar visualizations for the test regions can be found in Section 10.2. The smoothing method is based on the Savitzky-Golay filter [28]. As expected, NDVI of croplands increases sharply in spring, according to photosynthetic activity. Then, the NDVI values plateau, and they drop during autumn as for generic vegetation. The profile of non-cropland also follows a similar trend but with higher variability, this can be attributed to the photosynthetic activity of other types of vegetation such as grassland and forest. However, overall there is a clear gap between croplands and non-croplands across the year, supporting the separability of both spectral classes.

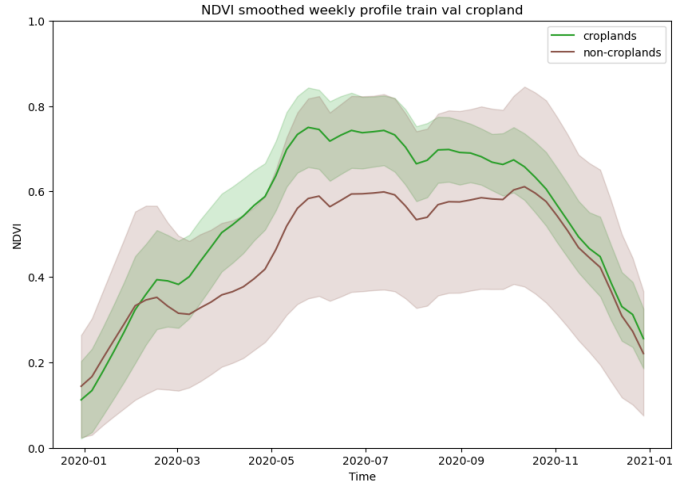

*Figure 6. Smoothed average NDVI weekly profile for croplands and non-croplands polygons on the training and validation region (missing data on the time series has been previously filled with linear interpolation). The shadowed area represents one standard deviation.*

## 3.6 Training methodology

As previously introduced, we consider two popular ML classifiers, namely support vector machines (SVM) and random forests (RF). A summary for each model can be found in the Supplementary Material, Section 10.1. We decided to perform model selection using cross-validation, to maximize the amount of data that is used for training. The use of a regular random cross-validation strategy can overestimate performance in spatial data due to the effect of spatial autocorrelation (SAC) [29], as it can be a source of data leakage when a training and a validation sample are spatially close. Recent studies [29], [30] show that ignoring spatial dependencies can indeed cause misinterpretation, data leakage, and overfitting. To overcome SAC, and improve generalization ability, we use spatial k-fold cross-validation (SKCV) [30], with the implementation of the `spacv` Python package [31]. As illustrated in Figure 7, we first split the region into several equal-size blocks, randomly assign them to k=3 folds, and then map each polygon to their corresponding fold. If a given polygon intersects with a boundary between blocks, then we allocate it to the block where the polygon's centroid is closer. This whole methodology contributes to making training samples more distant to validation samples, thus reducing data leakage and overconfident cross-validation metrics. A possible extra step to further separate training and validation samples is to set a dead zone radius, deleting all polygons falling within a radius of the validation set, thus avoiding data leakage altogether [30]. However, the trade-off between prediction performance and data collection costs should be considered, and given our limited training data, we omit the dead zone radius. And additional analysis about SAC can be found in Section 10.4.

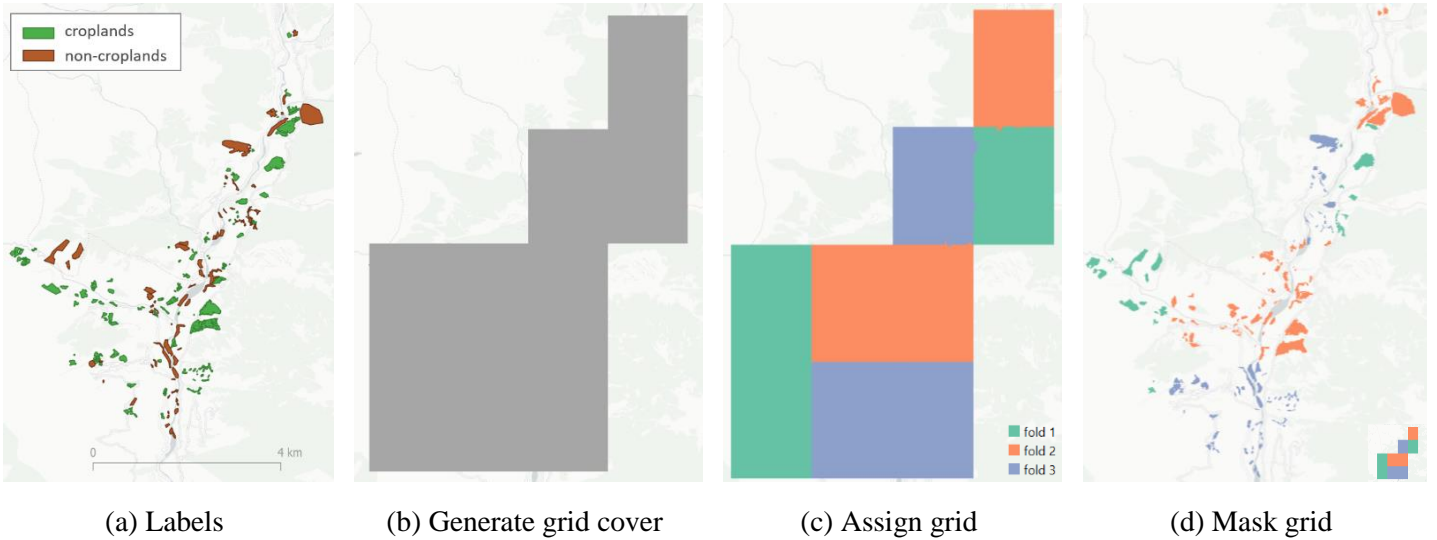

(a) Labels      (b) Generate grid cover      (c) Assign grid      (d) Mask grid

*Figure 7.  Steps to perform spatial cross-validation.*

Once all polygons are assigned to each of the three folds, regular cross-validation is performed. This means training on two folds at time, validating on the remaining one, and permuting three times. We use grid search to look for adequate hyper-parameters for each model, based on the search space defined in Table S1 of the Supplementary Material. We always compute accuracy, precision, recall and F1-score (i.e. harmonic mean between precision and recall) averaged over the three validation folds, but use accuracy to look for the best of the unique combinations of hyper-parameters.

# 4  Results

## 4.1  Cropland model selection

We run experiments for both models, following the training procedure described in Section 3.6, and the configurations about data preprocessing and feature engineering described in Sections 3.3 and 3.4, respectively. Table 3 summarizes the cross-validation results for the best RF and the SVM models, respectively. The selected hyper-parameters for each model are listed in Table S2 in Section 11.1 of the Supplementary Material. Both models exhibit good performance, but the SVM slightly outperforms the RF in all metrics. However, the RF is considerably faster than the SVM for making predictions over large-scale regions, therefore we chose it for generating the cropland map over our full study area.

*Table 3. Cross-validation scores of RF and SVM models for overall accuracy, precision, recall and F1-score.*

| Model / Metric | Accuracy | Precision | Recall | F1-score |
|---|---|---|---|---|
| *RF* | 0.862 | 0.905 | 0.827 | 0.860 |
| *SVM* | 0.878 | 0.926 | 0.840 | 0.878 |

## 4.2  Effects of feature selection

We performed an ablation experiment to study the effects of feature selection on the selected RF model. We focused on the group of spatial features only, in order to evaluate the importance of contextual information in a remote smallholding setting. This group of features is by far the largest, and NDVI-derived features are assumed to be always beneficial to tasks involving discrimination of land covers including vegetation. Three scenarios are compared: i) to add spatial features computed on all multi-spectral bands and vegetation indices (referred to as `all spatial`) as detailed in Section 3.4; ii) to add only the spatial features of the NDVI of surrounding pixels (`NDVI spatial`); and iii) do not consider any spatial information (`no`

`spatial`). Table 4 shows the cross-validation results for each metric in the three scenarios. Adding all the spatial features obtains the best cross-validation score on all metrics except recall. This indicates that including the contextual information is ultimately helpful for identifying the target pixel, but may cause the model to classify croplands as non-croplands compared to the other two cases, thus possibly underestimating the area of croplands.

*Table 4. Cross-validation scores of the selected RF model for overall accuracy, precision, recall and F1-score for an ablation study on spatial features.*

| Scenario / Metric | Accuracy | Precision | Recall | F1-score |
|---|---|---|---|---|
| *No spatial* | 0.854 | 0.860 | 0.855 | 0.855 |
| *NDVI spatial* | 0.852 | 0.861 | 0.858 | 0.854 |
| *All spatial* | 0.862 | 0.905 | 0.827 | 0.860 |

### 4.3 Feature importance evaluation

We inspect the feature importance of the selected RF model by the technique of *permutation feature importance*. The permutation feature importance is measured for a RF as the decrease in its accuracy when a single feature value is randomly shuffled [32]. We compute the permutation feature importance on the aggregation of the three test sets so that the results can highlight which features contributes the most to the generalization capability of the inspected model. Figure 8 shows the permutation feature importance scores that are greater than 0.001. We can see that the most important feature for the trained random forest model is the NIR mean value of neighboring pixels at the 24[th] week in 2020 (i.e. middle of summer), with a considerable margin over the other features. Given that leaves show strong reflectance at NIR wavelengths, especially during summer, the surrounding vegetation (or the lack of it) seems to be a discriminative factor. Overall, the NIR band and NDVI spatial features (i.e. the NIR and NDVI mean and standard deviation of the surrounding pixels) are the ones of most relevance for the selected model.

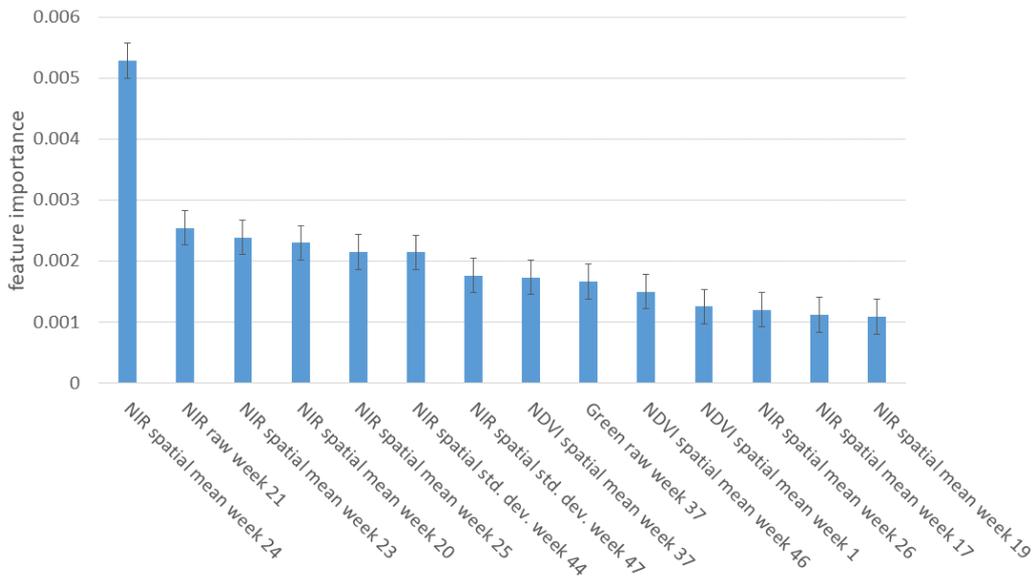

*Figure 8. Permutation feature importance of final random forest model on the three test sets. The error bars represents one standard deviation.*

## 4.4 Cropland model performance

We evaluated the performance of the selected cropland model by checking its accuracy, precision, recall, and F1-score, on the three test areas. All the test data has been prepared and preprocessed exactly as the training data (see Figure 2) and the respective full feature vector is computed and given to the RF model.

Table 5 shows the results obtained on all metrics for the three test datasets, including their weighted average (i.e. by pixel count). The test results in the Kullu district, exhibits a better performance compared to the other districts, Mandi and Shimla. This can be attributed to the geographical closeness to the training data and can be noted as well given the worse results of Shimla compared to Mandi, which lies closer to Kullu's training region. Nevertheless, the final model reaches an average overall accuracy and F1-score of approximately 87% for the three test sets, which is considered acceptable for the given application [33].

*Table 5. Test results for the selected RF model on the three districts for all metrics including their weighted average by pixel count.*

| District / Metric | Accuracy | Precision | Recall | F1-score |
|---|---|---|---|---|
| *Kullu* | 0.91 | 0.91 | 0.9 | 0.915 |
| *Mandi* | 0.87 | 0.87 | 0.89 | 0.875 |
| *Shimla* | 0.81 | 0.85 | 0.805 | 0.805 |
| *Average* | 0.869 | 0.884 | 0.863 | 0.870 |

## 4.5 Comparison of cropland model with existing maps

We qualitatively compare our RF predictions with two other open-source maps, namely the Copernicus land cover map [1] and GFSAD cropland map [2]. As mentioned, Copernicus provides a land cover type of cropland, while the GFSAD specifically detects croplands. Both of these benchmarks are generated by machine learning models, including SVM, RF and expert rules.

We show the cropland masks over Google satellite base map for several example locations in our test sets in Figure 9, where the white-masked area contains cropland predictions. The three sites have different vegetation levels, and the croplands are mixed with forests, grasslands, buildings, etc. All three maps do not perfectly match in their croplands prediction, however, our method provides more thorough predictions at a much higher spatial resolution (10-meter) than Copernicus (100-meter) and GFSAD (30-meter).

| Google base map | Ground truth | Our prediction | Copernicus[1] | GFSAD[2] |
|---|---|---|---|---|

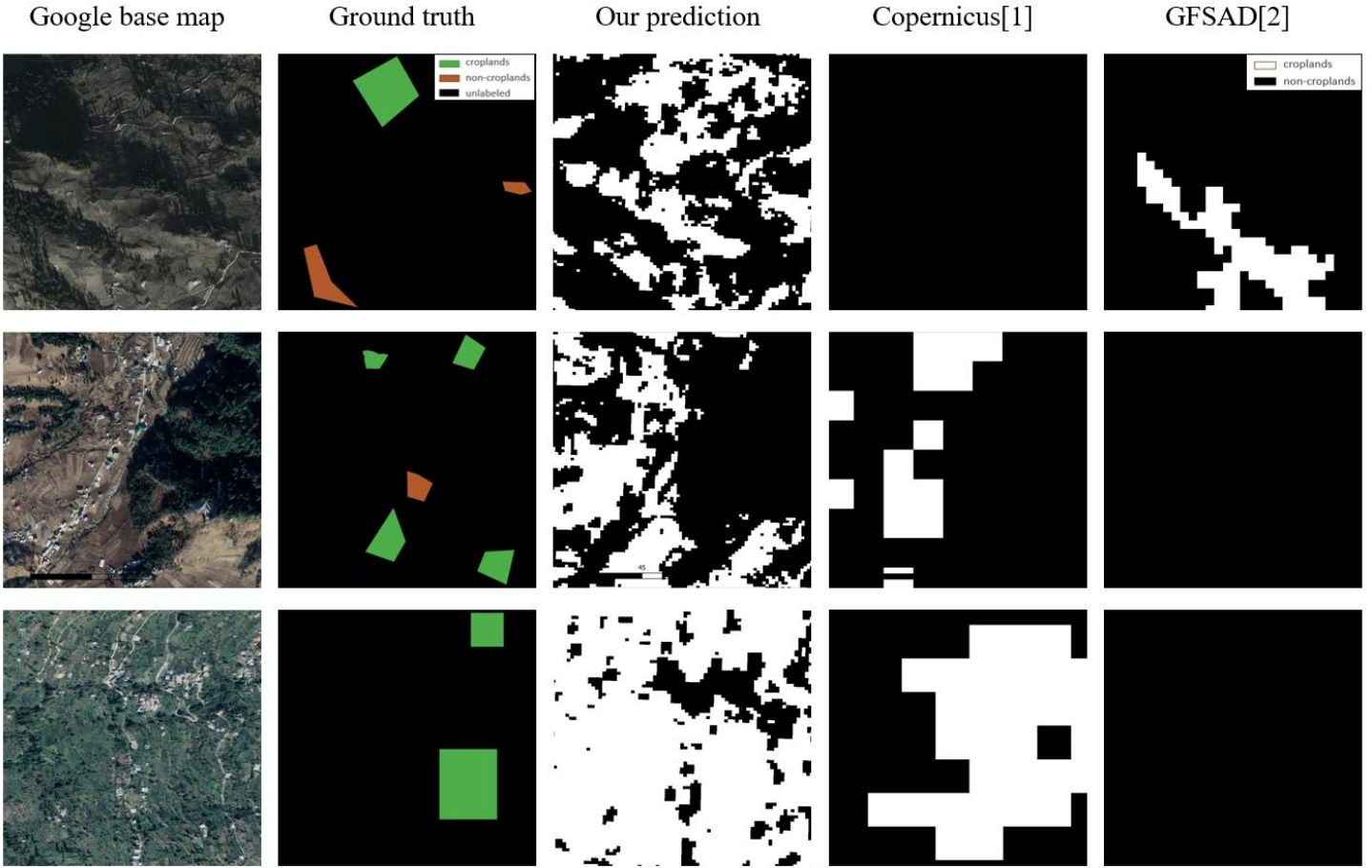

*Figure 9. Qualitative comparison of cropland map produced using our RF model, Copernicus land cover map [1], and GFSAD cropland map [2] for three example regions in our test sets. The Google base map image (2016 Dec. 30) is included for reference. We additionally include the ground truth polygons that have been labelled at these locations (note that there are not necessarily comprehensive). White pixels denote cropland predictions and black pixels non-cropland (except for the ground truth, where black denotes unlabeled).*

## 4.6 Open access cropland map of three districts in the state of Himachal Pradesh

Using our best RF model, we generate a cropland map in the full study region in the mountain regions of Himachal Pradesh. The output map is presented on Figure 10, spans over three districts and covers 14,600 square kilometers. We color the pixels by elevation, thus providing useful information to visually scan for different crops and species that grow at different altitudes. The elevation data is from the NASA Shuttle Radar Topographic Mission 90 m Digital Elevation Database [34], with a spatial resolution of about 90 meters and acquisition in 2018.

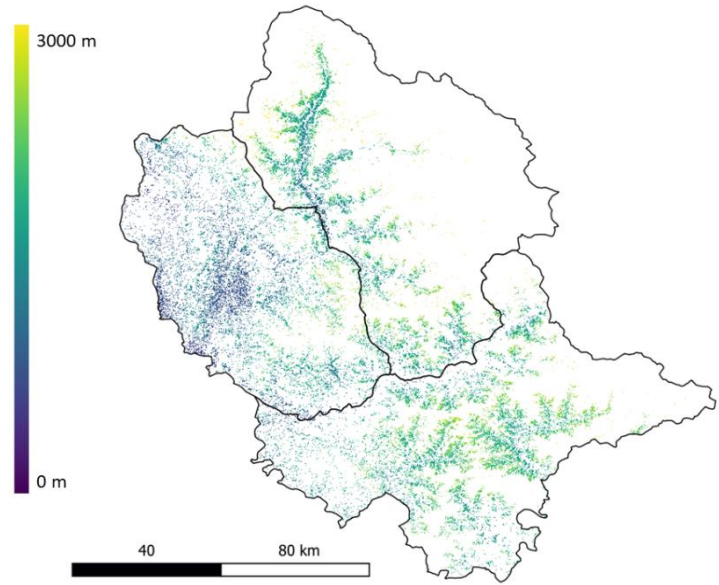

*Figure 10. Cropland prediction map of Kullu, Mandi, and Shimla districts in Himachal Pradesh, India, for the year 2020. The map is colored by elevation in order to facilitate the targeting of different crops, given their growing altitude range.*

## 5 Discussion

We developed a pixel-based, supervised machine learning model for cropland classification, based on the full time series of Sentinel-2 satellite images of 2020. With these high-resolution satellite images, we are able to obtain a 10-meter resolution map of cropland distribution. We validate our model on three test sets, where it can make accurate predictions (>87%) at a higher spatial resolution than two existing openly accessible cropland maps.

The produced cropland map presents good spatial generalization in the districts of our study area, regardless that the model was only trained on data of one district. These results could be attributed to both the handling of the spatial autocorrelation and the different types of features added during features engineering, that allow to capture relevant temporal and spatial patterns in the data. These features combined with global data normalization allow good generalization over areas not seen by the classifier at training time. However, croplands in drier regions with less vegetation might not always be successfully identified, as examples of those specific areas were not well covered in the training data. We visually examined some areas misclassified by our model and found that sometimes our model also regards wastelands with vegetation or forests as croplands. This could be attributed to deciduous forest being confused with fruit orchards such as apples, peaches or apricots, which are common plantations in the area. Additionally, we could notice that the model does not generalize well to predict cropland in flatter and lower altitude regions of the state. This can be expected, as our ground truth data was located in the hillier regions of Himachal Pradesh. However, we duly note that our training data is not representative to train a model able to generalize over the whole state. In this case, training data should be sampled from across the whole state or models pre-trained on available larger cropland datasets, such as the crowd-sourced Geo-Wiki dataset, could be used [27].

A potential improvement to the final map could be to include post-processing techniques in the pipeline, in order to reduce the salt and pepper effect (i.e. isolated predicted pixels), given that the resulting predictions present sparsely occurring positive and negative pixels in some regions. Some studies address this issue by segmenting neighboring pixels into homogeneous units while learning a pixel-based classifier to use objects as the unit for spatial analysis [9], [35]. Object-based classification has been observed to outperform pixel-based methods in some applications [36], [37], however the work of [38] shows no statistically significant advantage of using object-based image analysis for agricultural landscape classification in particular. Nevertheless, many studies have demonstrated the ability of object-based segmentation to improve spatial consistency of produced maps and reduce the salt-and-pepper effect from pixel-based classification [35], [37]. As a result, combining an object-based segmentation model into our pipeline for post-processing could potentially improve the visual quality of the produced map, e.g. by employing the well-stablished Markov Random Field and Conditional Random Field methods [39].

Finally, we limited ourselves to use only the 10 m resolution spectral bands of Sentinel-2 images (i.e. the RGB and NIR bands), following other studies on the subject [4], [35]. However, some studies do confirm the value of upsampling and including the 20 m resolution (i.e. red-edge and SWIR bands) spectral bands for vegetation classification [20], [40], and thus could be further explored in following work.

# 6 Conclusions

We successfully built a classification pipeline using machine learning to generate 10 m resolution cropland masks from time series of Sentinel-2 satellite images. We used it to create a binary map for the year 2020 over the districts of Kullu, Mandi and Shimla in Himachal Pradesh, India. We tackled the challenge of limited reliable ground truth data, which is often the case in remote regions in developing countries, and also of spatial autocorrelation, by adopting a spatial cross-validation approach to model selection. We evaluated support vector machine (SVM) and random forest (RF) classifiers, and selected the random forest as a final model given a good accuracy, comparable to SVM, but considerably lower inference time. We additionally performed an ablation study on our feature engineering and also analyzed the feature importance, confirming the relevance of the contextual features in explaining the model's performance for crop classification. We validated the model with three test sets, one on each district of the study area. Overall, our proposed RF approach achieved over 87% performance on accuracy, recall, precision and F1-score's weighted averaged over the three test datasets, demonstrating the model appropriateness for cropland identification in this region. Finally, we present our final cropland map visualizing the elevation of the predicted cropland pixels, in order to facilitate its use when targeting for specific crop species by inspecting for their usual growing altitude range. Our cropland map has a higher spatial resolution (10-meter) than existing cropland maps, such as the Copernicus land cover map (100-meter) and GFSAD (30-meter) cropland map. By means of a qualitative assessment with the aforementioned maps, we show that we provide visually more accurate cropland maps, while having higher spatial accuracy. This is unsurprising, since our pipeline is specialized for the task and trained on local data. Our map is made openly accessible via Google Earth Engine for agricultural organizations, governments, farmer producer groups, and supply-chain stakeholders.

# 7 Acknowledgments

This work was funded by the data.org Inclusive Growth and Recovery Challenge grant "Your Virtual Cold Chain Assistant", supported by The Rockefeller Foundation and the Mastercard Center for Inclusive Growth. The funder was not involved in the study design, collection, analysis, and interpretation of data, the writing of this article, or the decision to submit it for publication. We would like to thank Kendall Nowocin, Srinivas K. Marella, and their team in CoolCrop for their support with data collection and for providing valuable insights from the Indian horticulture sector, which helped us in improving this work. We would also like to thank the "Your Virtual Cold Chain Assistant" team for valuable discussions and feedback. This manuscript has been released as a preprint at arXiv.

# 8 Author contributions

D.L developed the code, and wrote the first draft of the paper. J.G supervised D.L during her internship and together developed the methodology. J.G, M.V and T.D critically reviewed and refined the paper. T.D and J.G acquired the funding for the project on which this work relied.

# 10  Supplementary material

## 10.1  Selected models

We employed two supervised models for cropland classification, from the `scikit-learn` package implementation [41]. Some details about the models are outlined in this section.

### 1)  Support Vector Machines (SVM)

Support Vector Machines (SVMs) are among the most popular supervised learning algorithms solving both regression and classification problems. The objective of SVM is to find a hyperplane that separates data points of one class from the other by learning a largest possible separating hyperplane between the two classes. At the same time, a certain number of misclassification is allowed, controlled by a penalty term in the objective function. To efficiently perform a non-linear classification, a kernel function is defined over input features, implicitly mapping them into a higher-dimensional space where a linear separation is likely to be more effective. Choosing a correct kernel function and an optimal regularization term, which control the width of the hyperplane, are important steps to avoid overfitting. An illustration of this model is presented in Figure S1.

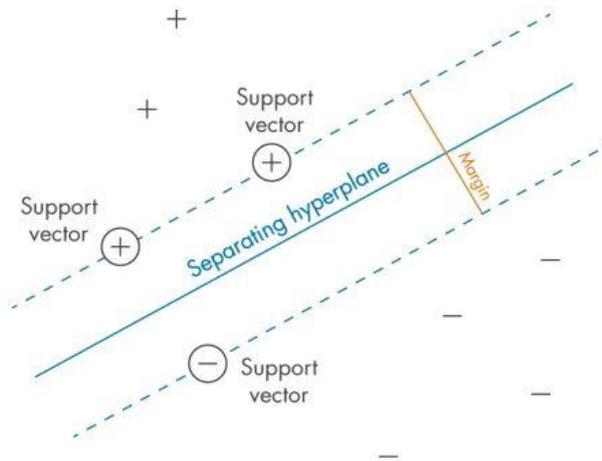

*Figure S1.  An illustration of support vector machines with a hard margin. Plus and minus marks represent two different classes. (Source: https://www.mathworks.com/discovery/support-vector-machine.html)*

### 2)  Random Forest (RF)

Random Forests (RFs) are an ensemble learning method for classification and regression. It fits several decision tree classifiers on various random subsets of the training dataset and establishes outputs based on predictions of each tree by majority voting or taking the average. Each decision tree in the ensemble is built by binary partitioning algorithms, and the final tree structure (i.e. depth and leaf size) is determined after all training samples in a given training split are fed to the tree and each split determined. The latter are thresholds on feature vectors representing a given data bag, coming either from the root of the tree (the whole subset) or from another node. As the name of the model suggests, the selection of the features is made on a random subset of features, and the one showing best split score (e.g. Gini score or lowest entropy) is kept. A main advantage of random forest, is that each split, each tree, and therefore the whole forest is invariant to feature scaling. Aggregating multiple trees benefit from collective wisdom, making the model more robust to overfitting and often achieving higher accuracy and better estimates than single decision tree, which are extremely prone to overfit the training data. Still, it also comes at the expense of a slight increase in the bias and some sacrifice of interpretability. Moreover, RF can parallelize the construction of trees, since each tree is built independently from each other. An illustration of this model is presented in Figure S2.

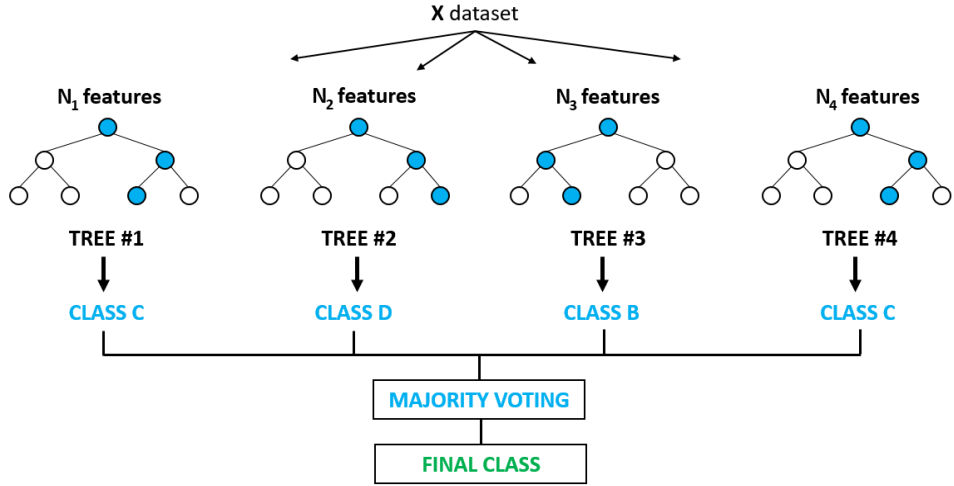

*Figure S2. An illustration of a random forest [42].*

### 3) Hyper-parameters

Each of the previous models have specific hyper-parameters (i.e. non-trainable parameters). We selected a search space for performing grid search during cross-validation of each model, as defined in Table S1. The hyper-parameters configuration yielding the most performant models during cross-validation, and that are reported in Section 4.1, are listed in Table S2.

*Table S1. A summary of hyper-parameters and their searching space in our experiments. C is a regularization parameter from SVMs and gamma is a kernel coefficient. The number of estimators for RF are the number of trees, the criterion is the function used to measure the quality of a split, the maximum depth is the maximum depth of any tree, and the maximum number of samples is the sub-sample size for bootstrapping.*

| Models | Hyper-parameters and their searching space |
|---|---|
| *SVM* | C=[0,5, 1, 10, 100]; kernel=[poly, rbf]; gamma=[scale] |
| *RF* | n_estimators=[100, 300, 500]; criterion=[gini, entropy]; max_depth=[5, 10, 15]; max_samples=[0.5, 0.8, 1] |

*Table S2. A summary of the hyper-parameter configuration of the selected models during cross-validation.*

| Models | Best hyper-parameters configurations |
|---|---|
| *SVM* | C=[0,5]; kernel=[poly]; gamma=[scale] |
| *RF* | n_estimators=[100]; criterion=[entropy]; max_depth=[15]; max_samples=[0.5] |

## 10.2 NDVI profile visualization

We checked the label quality in all training and testing regions by NDVI profile as suggested in Section 3.5. Figure S3 shows the profiles for each region. As a reminder the NDVI, is defined as the ratio between the difference and the sum of the NIR and red bands, and it serves as a proxy for photosynthetic activity. We can see that NDVI of croplands in all datasets has a similar pattern, sharply increasing in spring, then entering a plateau and dropping during autumn. There are also clear gaps between croplands and non-croplands in Kullu training and testing regions, and Shimla test regions, but it is not that obvious in Mandi, probably due to the small number of labels in that region.

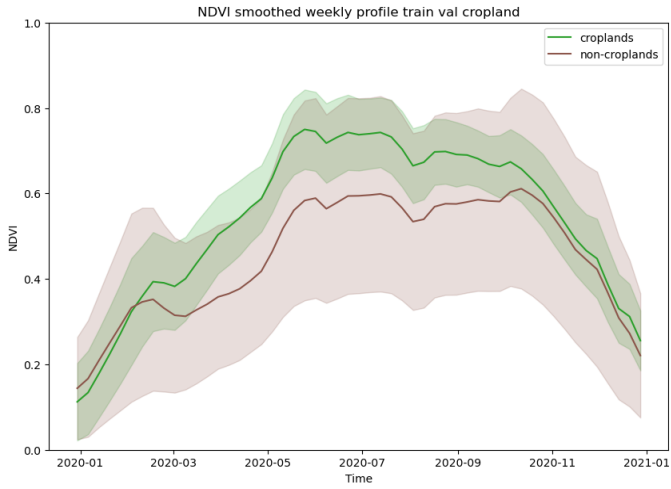

(a) NDVI profile of Kullu training region.

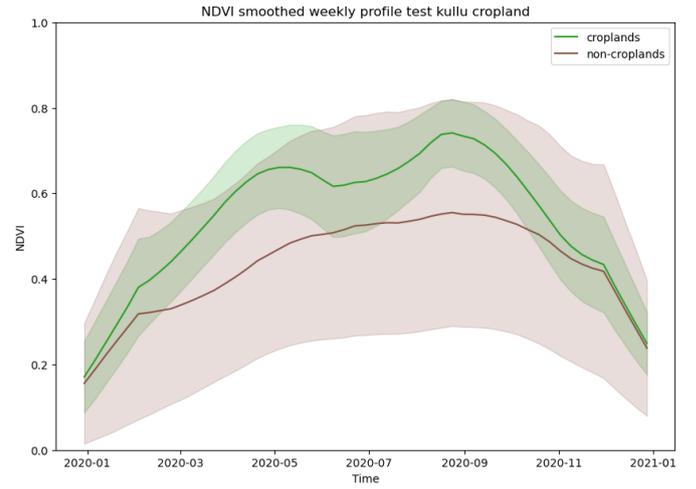

(b) NDVI profile of Kullu test region.

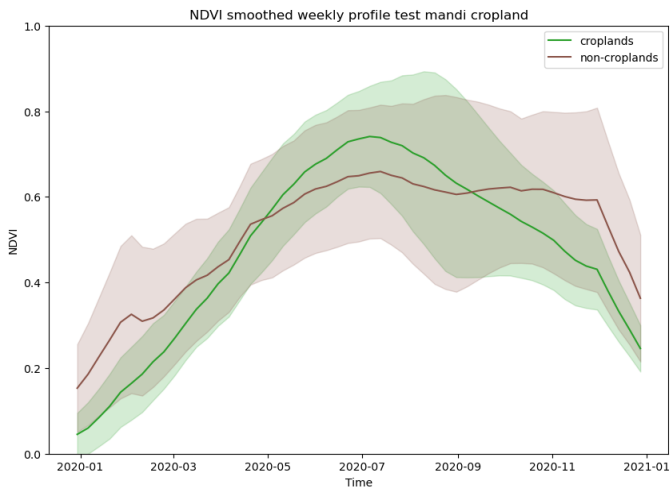

(c) NDVI profile of Mandi test region.

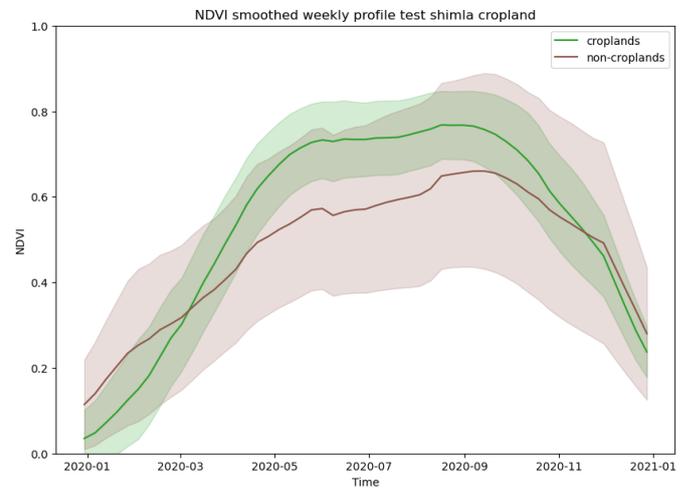

(d) NDVI profile of Shimla test region.

*Figure S3. Smoothed NDVI weekly profile for croplands and non-croplands using different datasets (with linear interpolation for missing data). Shadows represent standard deviation.*

## 10.3 Effects of data normalization and imputation methods

We investigated the effects of different methods for handling the missing data and for normalizing the data. Different models have different sensitivity to data normalization methods. We implement nonlinear SVM using a Gaussian Radial Basis Function kernel, which is an exponential of a scaled Euclidean distance between data points. For the Euclidean distance to be sensitive to every feature considered, they should be scaled in a way that all of the features contribute to the distance measure. In contrast, only relative order matters for RF, so it is invariant to the data normalization method used.

The handling of missing data influences how biased a dataset could be. We compare the forward filling and linear interpolation methods. As shown in Figure S4, linear interpolation is always better than forward filling in all models and with all scaling methods. This is expected as linear interpolation balances information from previous and following images in time, while forward filling uses less information. The differences between the normalization methods are smaller than those of data imputation methods. The preferred choice here is converting data into reflectance values, by multiplying by 10000. The other methods considering per feature scaling might distort the relationships of the data in the temporal dimension. In addition, this method gives results that make physical sense, which may be an advantage compared to the image `as_float` method.

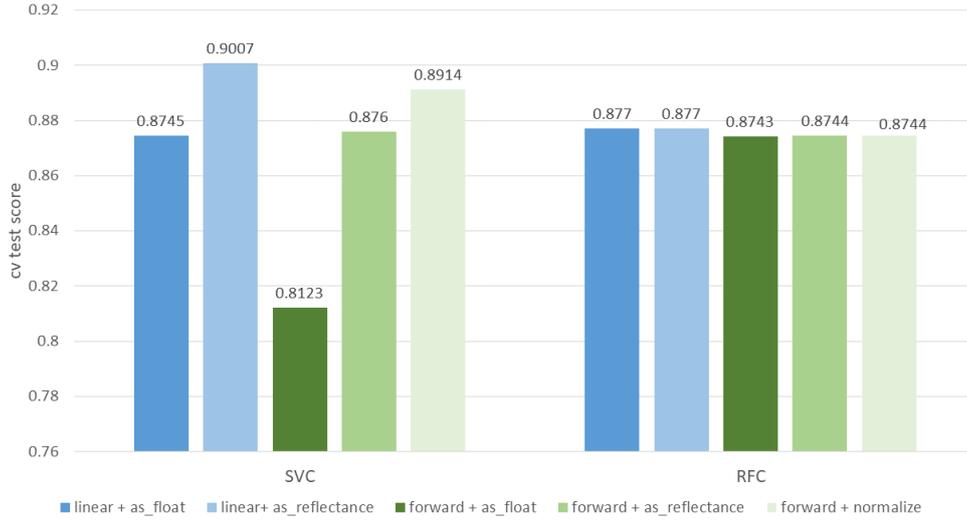

*Figure S4. Cross-validation scores of two studied models with different combinations of data normalization methods detailed in Section 3.3 (as_float, as_reflectance, normalize) and imputations methods for missing values (linear interpolation, forward filling).*

### 10.4 Spatial autocorrelation quantification

The semivariograms are one of the most common ways to confirm the presence of spatial autocorrelation (SAC) in the data [30], [43]. It is defined as half the average squared difference between the values at two points separated at a certain distance (lag) [44]. Formally, it can be defined as: $\gamma(s_i, s_j) = \frac{1}{2} Var(Z(s_i) - Z(s_j))$, where $Z(s)$ denotes the value at spatial coordinates $s = (x, y)$ and $Var(\cdot)$ is the variance. Hence, the semivariogram $\gamma(s_i, s_j)$ can be seen as a dissimilarity function. If SAC affects the dataset, the semivariogram curve should increase smoothly as a function of the distance. A term to describe the distance at which the semivariogram levels off to a flat line is called the range. The range can be interpreted as the distance at which autocorrelation stops to be important.

We plot the experimental semivariogram in Figure S5 below with a sample of our training data (i.e. taking one every 2000 pixels). Because of the scale of all points, the number of closer distances cannot be shown clearly. Still, small autocorrelation at about 2.5 km can be observed. A range in this case is between 3 and 5 km, therefore the validation data should be kept separated from the training data to this distance. Note that given the closeness of our training data, as illustrated in Figure 7, SAC cannot be fully eradicated since the polygons at the edges of each fold will sometimes be unavoidably close to validation polygons of a neighboring fold.

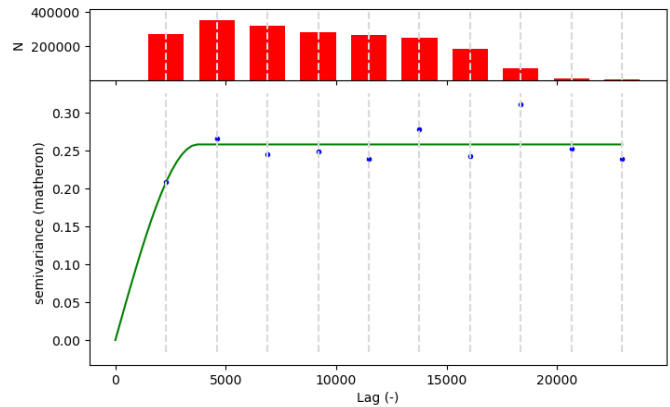

*Figure S5. The semivariogram of 2000 randomly sampled training pixels. The upper red bar represents the occurrence of the corresponding pairwise distance in samples. The blue dots are calculated semivariogram at such a lag, and the green curve fits the blue dots. Spherical model is the theoretical variogram function to be used to describe the experimental variogram.*